\documentclass[10pt,twocolumn,letterpaper]{article}

\usepackage{cvpr}              % To produce the CAMERA-READY version
\usepackage{graphicx}
\usepackage{amsmath}
\usepackage{amssymb}
\usepackage{booktabs}
\usepackage{microtype}
\usepackage{multirow}
\usepackage{amsmath}
\usepackage{amssymb}
\usepackage{array}
\usepackage{enumitem}
\usepackage{booktabs}
\usepackage{bbding}
\usepackage[switch]{lineno}
\usepackage[symbol]{footmisc}

\usepackage[pagebackref,breaklinks,colorlinks]{hyperref}

\usepackage[capitalize]{cleveref}
\crefname{section}{Sec.}{Secs.}
\Crefname{section}{Section}{Sections}
\Crefname{table}{Table}{Tables}
\crefname{table}{Tab.}{Tabs.}

\def\cvprPaperID{1845} % *** Enter the CVPR Paper ID here
\def\confName{CVPR}
\def\confYear{2022}

\begin{document}

\title{Learning to Refactor Action and Co-occurrence Features for \\ Temporal Action Localization}

%\author{Kun Xia, Le Wang, \\
%% omit the following lines up until the closing ``}''.
%% Additional authors and addresses can be added with ``\and'',
%% just like the second author.
%% To save space, use either the email address or home page, not both
%\and

\author{Kun Xia$^1$ ~Le Wang$^1$\footnote[1]{} ~~Sanping Zhou$^1$ ~Nanning Zheng$^1$ ~Wei Tang$^2$ \\
	$^{1}$Institute of Artificial Intelligence and Robotics, Xi'an Jiaotong University\\
	$^{2}$University of Illinois at Chicago\\
	{\tt\small xiakun@stu.xjtu.edu.cn; $\{$lewang, spzhou, nnzheng$\}$@xjtu.edu.cn; tangw@uic.edu}
}
\maketitle
\pagestyle{empty}  % no page number for the second and the later pages
\thispagestyle{empty} % no page number for the first page
\footnotetext{$^*$Corresponding author.}
%%%%%%%%% ABSTRACT
\begin{abstract}
   The main challenge of Temporal Action Localization is to retrieve subtle human actions from various co-occurring ingredients, \textit{e.g.}, context and background, in an untrimmed video.
   While prior approaches have achieved substantial progress through devising advanced action detectors, they still suffer from these co-occurring ingredients which often dominate the actual action content in videos.
   In this paper, we explore two orthogonal but complementary aspects of a video snippet, \textit{i.e.}, the action features and the co-occurrence features.
   Especially, we develop a novel auxiliary task by decoupling these two types of features within a video snippet and recombining them to generate a new feature representation with more salient action information  for accurate action localization.
   We term our method \emph{RefactorNet}, which first explicitly factorizes the action content and regularizes its co-occurrence features, and then synthesizes a new action-dominated video representation.
   Extensive experimental results and ablation studies on THUMOS14 and ActivityNet v1.3 demonstrate that our new representation, combined with a simple action detector, can significantly improve the action localization performance.
\end{abstract}

%%%%%%%%% BODY TEXT
\section{Introduction}
\label{sec:intro}

Temporal Action Localization~(TAL) aims to locate the start and end times of action instances from long untrimmed videos as well as to classify their categories. As a fundamental task in video understanding, it has attracted great attention in recent years and facilitated various applications such as security surveillance~\cite{simonyan2014two, yuan2017temporal, wray2021semantic} and human behavior analysis~\cite{fan2018end, zhai2020two, tirupattur2021modeling}.
\begin{figure}[t]
	\centering
	\includegraphics[width=\linewidth]{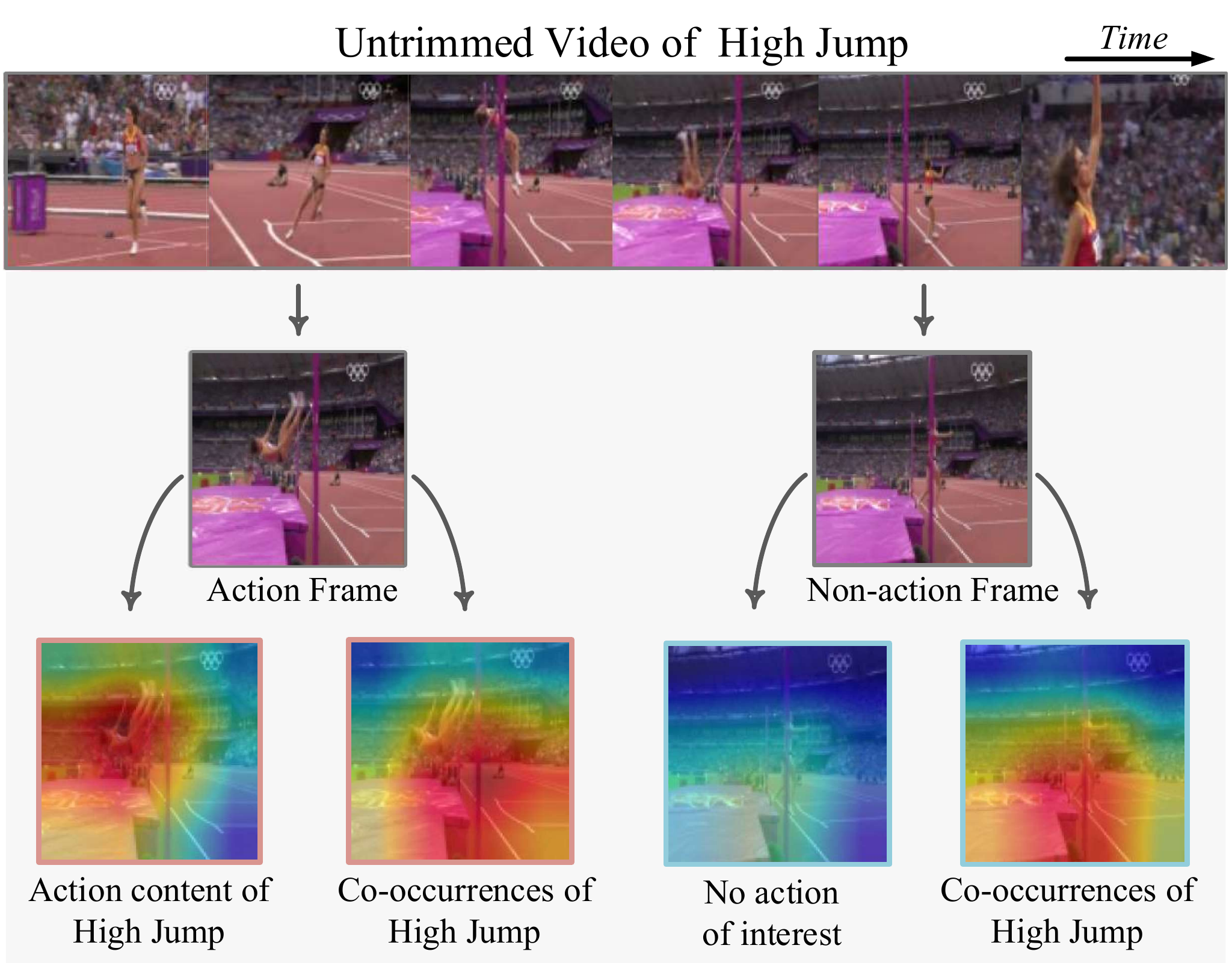}
	\caption{Visualization of the action component and the co-occurrence component decoupled by our method from a snippet representation. 
	Co-occurrence components help reduce action ambiguity and uncertainty, but they often dominate the action components in a video, and thus adversely affect action detection. How to balance these two components in a snippet representation is an important yet under-explored problem.
	%A promising solution of temporal action localization is to learn how to take the best of both features to locate temporal actions in long untrimmed videos.
	}
	\label{fig:motivation}
\end{figure}

A common first step of current TAL approaches~\cite{gong2020scale, xu2020g, zeng2021graph, qing2021temporal} is to extract features from each video snippet via a pre-trained two-stream network. They are then aggregated, \textit{e.g.}, via max pooling, to obtain the representation of a proposal or an anchor for action classification and temporal boundary regression. These snippet-level features characterize two aspects of a video snippet, and we term them the \textit{action component} and the \textit{co-occurrence component}, respectively.
The \textbf{action component} refers to features characterizing the action occurring in a snippet, including the motion pattern of one or more persons and their interaction with objects. 
The \textbf{co-occurrence component} refers to features not characterizing any actions but often co-occurring with them in a frame or a snippet. This includes class-specific \textit{context}, which only co-occurs frequently with certain actions, \textit{e.g.}, a track field, and class-agnostic  \textit{background}, whose occurrence is less relevant to the action categories, \textit{e.g.}, the sky.
Figure~\ref{fig:motivation} illustrates that an untrimmed video of \emph{High Jump} contains a subtle action component and a richer co-occurrence component surrounding the action.

It is critical to treat both the action and co-occurrence components carefully to achieve robust TAL. 
On the one hand, while the action component directly characterizes an action, it can be ambiguous and uncertain because the motion and appearance of humans and their interactions with each other and with objects are complex. 
Thus, only retaining the action component from the snippet-level features is insufficient. On the other hand, while some co-occurrence components are useful to reduce the action ambiguity and uncertainty, \textit{e.g.}, a swimming pool distinguishing ``diving'' from ``trampoline'', some others are more like nuisance noise, \textit{e.g.}, audience and random persons in the scene, and over-relying on the co-occurrence component also blurs the action boundaries. Therefore, it is necessary to find an appropriate balance between the action component and the co-occurrence component  in a feature representation, especially as the latter often dominates the former in a video. This problem has been largely ignored by prior work.

This paper investigates \textit{feature refactoring} for TAL. It means to \emph{decouple} the snippet-level features as the action component and the co-occurrence component, and then \emph{recombine} them into a more appropriate representation to achieve effective TAL. 

We propose a novel \textit{Feature Refactoring Network} or \textit{RefactorNet} to achieve this goal. It consists of a feature decoupling module and a feature recombining module. Since the annotations of co-occurrence components are unavailable, we collect action samples and their \textit{coupling samples} from the whole video for the decoupling process. Coupling samples refer to any video snippets containing co-occurring elements of the action but do not involve the actual action. In other words, an action sample and its coupling sample share similar co-occurrence components but differ in whether the action component is present. Taking the action samples and coupling samples as supervision, the feature decoupling module is trained to separate action and co-occurrence components. Then, the feature recombining module synthesizes these two components into a new snippet representation containing a more salient action component and a more suitable co-occurrence component for accurate action detection.
%This paper introduces \emph{feature refactoring} for TAL. It means to decouple the snippet-level features as the action component and the co-occurrence component, and then recombine them into a more appropriate representation to achieve effective TAL. We propose a novel Feature Refactoring Network or \textit{RefactorNet} to achieve this goal. 
%\textcolor{blue}{It consists of two encoders. The two encoders are trained to decouple the snippet feature into the action component and the co-occurrence component, respectively. Then, these two components are synthesized a new snippet representation.}
%
%Since the annotations of co-occurrence components are unavailable, we collect action samples and their \emph{coupling samples} from the whole video for the decoupling process. Coupling samples refer to any video snippets containing co-occurring elements of the action but do not involve the actual action. In other words, an action sample and its coupling sample share similar co-occurrence components but differ in whether the action component is present. 
%Then, we take action samples and coupling samples as supervision and introduce two cosine similarity losses to encourage the two encoders to separate action and co-occurrence components. Furthermore, a KL divergence loss is used to regularize the distribution of the co-occurrence component. Finally, the new snippet representation contains a more salient action component and a more suitable co-occurrence component for accurate action detection.

Both quantitative and qualitative results demonstrate that our RefactorNet can effectively decouple the action and co-occurrence components, and the recombined snippet representations improve both action classification and temporal boundary regression.
Combined with a simple action detector, our RefactorNet achieves state-of-the-art performance on two benchmarks, THUMOS14 and ActivityNet v1.3. 

%verify the effectiveness of the proposed method.

Our contributions can be summarized as follows:
\begin{itemize}
%	\item We tackle a relatively under-explored problem of refactoring action and co-occurrence components for TAL, which is notable but largely neglected by prior work.
%	\item We present a novel \textit{RefactorNet} that effectively decouples and  recombines action and co-occurrence components for accurate action localization.
	\item Co-occurrence components help reduce action ambiguity and uncertainty, but they often dominate the action components in a video, and thus adversely affect TAL. How to balance these two components in a snippet representation is an important yet under-explored problem. Our proposed RefactorNet is the first approach to explicitly refactor, \textit{i.e.}, decouple and recombine, these two components to obtain a new snippet representation containing a more salient action	component and a more suitable co-occurrence component for accurate action detection.
	\item Decoupling the two components is indeed very challenging because they co-occur frequently, and more severely, their annotations are unavailable. To address these difficulties, we carefully design the learning objective, and introduce action samples and their coupling samples, which can be obtained from standard TAL annotations, to supervise the decoupling process. They together help our feature decoupling module separate the two components effectively.
	\item Our RefactorNet outperforms all state-of-the-art methods on two benchmark datasets.
	Extensive ablation study and visualizations are provided to show in-depth analysis of the decoupling process and validate how it improves TAL.
\end{itemize}

\begin{figure*}[t]
	\includegraphics[width=\linewidth]{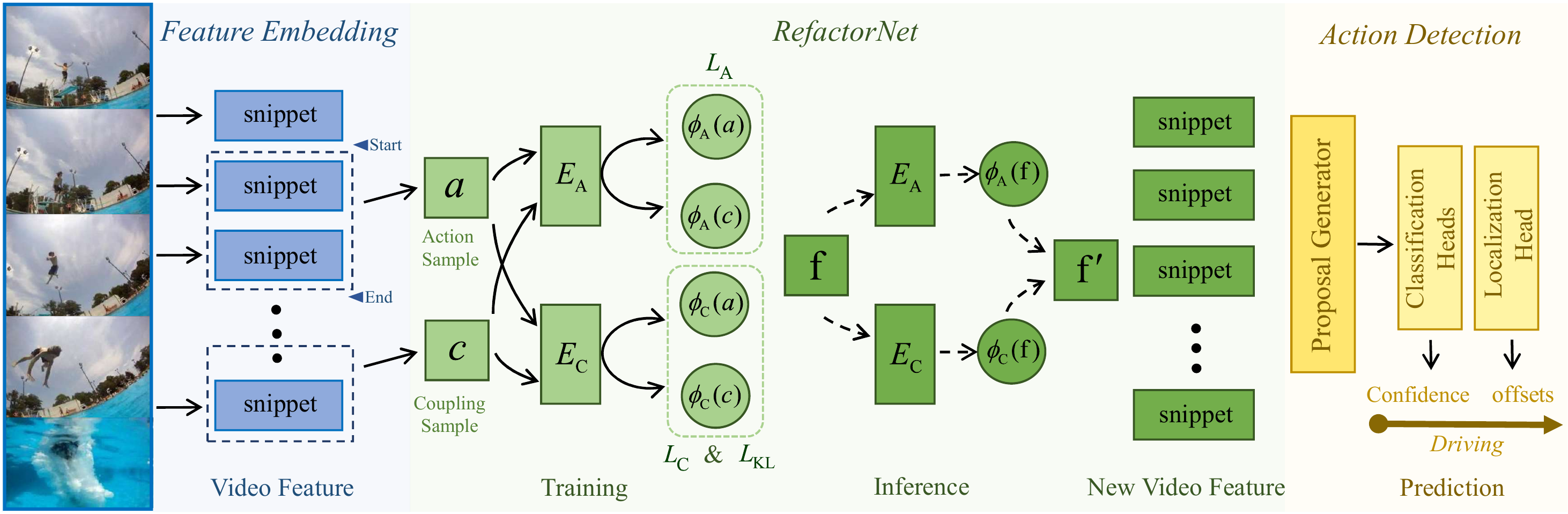}
	\caption{The overall pipeline of the proposed framework. It consists of three stages, \textit{i.e.}, \emph{Feature Embedding}, \emph{RefactorNet}, and \emph{Action Detection}. First, the feature embedding stage extracts original snippet features through a two-stream network. Subsequently, our proposed RefactorNet is trained to decouple a snippet representation into an action component and its co-occurrence component, supervised by pairs of action samples and coupling samples. During the inference process, each snippet feature vector~$\textbf{f}$ is decoupled into an action component~$\boldsymbol{\phi}_{\mathrm{A}}(\textbf{f})$ and a co-occurrence component~$\boldsymbol{\phi}_{\mathrm{C}}(\textbf{f})$, which are then recombined to generate a new snippet feature vector~$\textbf{f}^{\prime}$. Finally, the action detector locates actions from the video via the new snippet representations.}
	\label{fig:pipeline}
\end{figure*}

%-------------------------------------------------------------------------
\section{Related Work}
\textbf{Temporal Action Localization.}
Recently, TAL has made substantial progress.
CDC~\cite{shou2017cdc} performs convolution and de-convolution on the input video for dense score prediction, and combines segment proposals to detect action instances.
BSN~\cite{lin2018bsn} densely predicts the start and end probabilities of actions to generate high-quality action proposals with precise boundaries.
Yang~\textit{et al.}~\cite{yang2018exploring} propose the TPC network to preserve the temporal resolution, enabling frame-level granularity action localization with minimal loss of temporal information.
GTAN~\cite{long2019gaussian} proposes a one-stage TAL framework to model action instances with various lengths through learning a set of Gaussian kernels.
AFSD~\cite{lin2021learning} presents a purely anchor-free TAL model to predict the temporal distance of the start and end from each location and the action class.
In addition, other works~\cite{lin2020fast, jain2020actionbytes, zhao2020bottom, gao2020accurate, yang2021background} have attractive ideas that advance the TAL field further.
Especially, some researchers realize that accurately inferring human actions requires understanding video context. 
To this end, they model relations between proposals or between snippets, \textit{e.g.}, P-GCN~\cite{zeng2019graph}, G-TAD~\cite{xu2020g}, TCANet~\cite{qing2021temporal}, ContextLoc \cite{zhu2021enriching}, to enhance their features.
In contrast, this paper addresses a different problem that the co-occurrence component within a video snippet often dominates the actual action, through explicitly decoupling and recombining these two components within each individual snippet.
Therefore, our feature refactoring is complementary to relation modeling \cite{zeng2019graph,xu2020g,qing2021temporal,zhu2021enriching}, which is also validated by our experiments.

% Therefore, refactoring action and co-occurrence components will facilitate detection performance and the effectiveness of context modeling through getting rid of over-relying on the co-occurrence component.

Other related works extract the action frames of interest from false positive frames in Weakly-Supervised Temporal Action Localization~(WSTAL), where only the video-level labels are available.
BaSNet~\cite{lee2020background} introduces an auxiliary background class to suppress the interference of background frames to actual action frames.
Liu~\textit{et al.}~\cite{liu2021weakly} divide video snippets into positives and negatives to learn action and context sub-spaces. 
Different with them, we do not introduce additional supervision for the co-occurrence component, as it may require detailed annotations and it is difficult to generalize to larger datasets. Moreover, we do not consider the co-occurrence component as negative samples, which often contain contextual information useful to action classification. 

\textbf{Disentangled Representation Learning.}
The objective of disentangled representation learning is to extract interesting features or synthesize new visual representation, which has developed significantly in recent years~\cite{denton2017unsupervised, yingzhen2018disentangled, choi2019can, pan2021videomoco, wang2021enhancing} and has been widely applied in various fields,~\textit{e.g.}, image deblurring~\cite{lu2019unsupervised}, face identification~\cite{liu2018exploring, wu2019disentangled} and semantic segmentation~\cite{liu2021unsupervised}, etc.
Villegas~\textit{et al.}~\cite{villegas2017decomposing} decompose the inputs for video prediction into motion and content. Thereafter, predicting the next frame reduces to converting the extracted content features into the next frame content by the identified motion features.
Hsieh~\textit{et al.}~\cite{hsieh2018learning} decompose the high-dimensional video into low-dimensional temporal dynamics to predict future video frames.
Wang~\textit{et al.}~\cite{wang2018pulling} exploit action samples and the corresponding conjugate samples to deliberately separate action from context  for human action recognition.
Eom~\textit{et al.}~\cite{eom2019learning} present a GAN to disentangle identity-related and -unrelated features using identification labels for person re-identification.
Hamaguchi~\textit{et al.}~\cite{hamaguchi2019rare} use disentangled representation learning to disentangle variant and invariant factors that represent trivial events and image contents for the rare event detection task.  
Bahng~\textit{et al.}~\cite{bahng2020learning} develop a novel algorithm, ReBias. It solves a min-max problem, where the target is to promote the independence between the network prediction and all biased predictions. 
Singh~\textit{et al.}~\cite{singh2020don} focus on addressing context biases for visual classifier by explicitly learning a robust feature subspace of a category.  
Huang~\textit{et al.}~\cite{huang2021self} design a self-supervised video representation learning method to decouple the context and motion representations from compressed videos.
Inspired by these literature, in this paper, we introduce disentangled representation learning into the field of TAL.

\section{RefactorNet}
The pipeline of our entire framework is presented in Figure~\ref{fig:pipeline}. The proposed RefactorNet is inserted between the feature embedding module and the action detection module to refactor the snippet feature representation.
We elaborate the proposed network and action detection framework below.

\subsection{Problem Setting}
%TAL is formulated as a multi-task learning problem, which includes two sub-tasks: action localization and action classification.
Given an untrimmed video, we extract the visual features of every a few consecutive frames as a snippet feature vector~$\mathbf{f}\in\mathbb{R}^C$, where $C$ is the feature dimension. Then, we can represent the video feature sequence as $\mathbf{F} \in \mathbb{R}^{C \times L}$, where $L$ is the number of snippets in the video. The ground truth annotation is a set of action instances $\Psi=\left\{\varphi_{n}=\left(t_{s, n}, t_{e, n}, c_n\right)\right\}_{n=1}^{N}$, where $t_{s,n}$, $t_{e,n}$ and $c_n$ represent the start time, end time and action class of an action instance $\varphi_{n}$, respectively. ${N}$ is the total number of action instances in the video.

\subsection{Motivation and Overview}
Prior methods perform TAL upon the original video features $\mathbf{F}$ extracted by a two-stream network.
However, the co-occurrence component, including context and background, within each video snippet often dominates the subtle action component.
%While some co-occurrence components are useful to reduce the action ambiguity and uncertainty, \textit{e.g.}, a swimming pool distinguishing ``diving'' from ``trampoline'', some others are more like nuisance noise, \textit{e.g.}, audience and random persons in the scene, and over-relying on the co-occurrence component makes the TAL model less sensitive to the actual action content.
While some co-occurrence components are useful to reduce the action ambiguity and uncertainty, a TAL model can over-rely on the co-occurrence component after training, and confuses it with the actual action content during testing.
% which causes the ambiguity of the actual action content, and confounds the detection algorithm due to over-relying on co-occurrence components.
We seek an appropriate balance between action and co-occurrence components, through amplifying the action signal within the video snippet and regularizing the co-occurrence component, to improve action localization. 

To this end, we propose a novel RefactorNet, as illustrated in Figure~\ref{fig:pipeline}. It aims at synthesizing new snippet features through decoupling and recombining the action component and the co-occurrence component based on action samples and coupling samples. Finally, the new feature representation is used for action localization.
Specifically, our approach consists of three steps. 
(1) Collect action samples and their coupling samples. % that entangled but do not contain actual actions across the video. 
(2) Explicitly decouple action and co-occurrence components based on pairs of action samples and coupling samples. (3) Effectively recombine these two components and synthesize new snippet features for video action detection.

\subsection{Collecting action samples and coupling samples}
\begin{figure}[t]
	\centering
	\includegraphics[width=\linewidth]{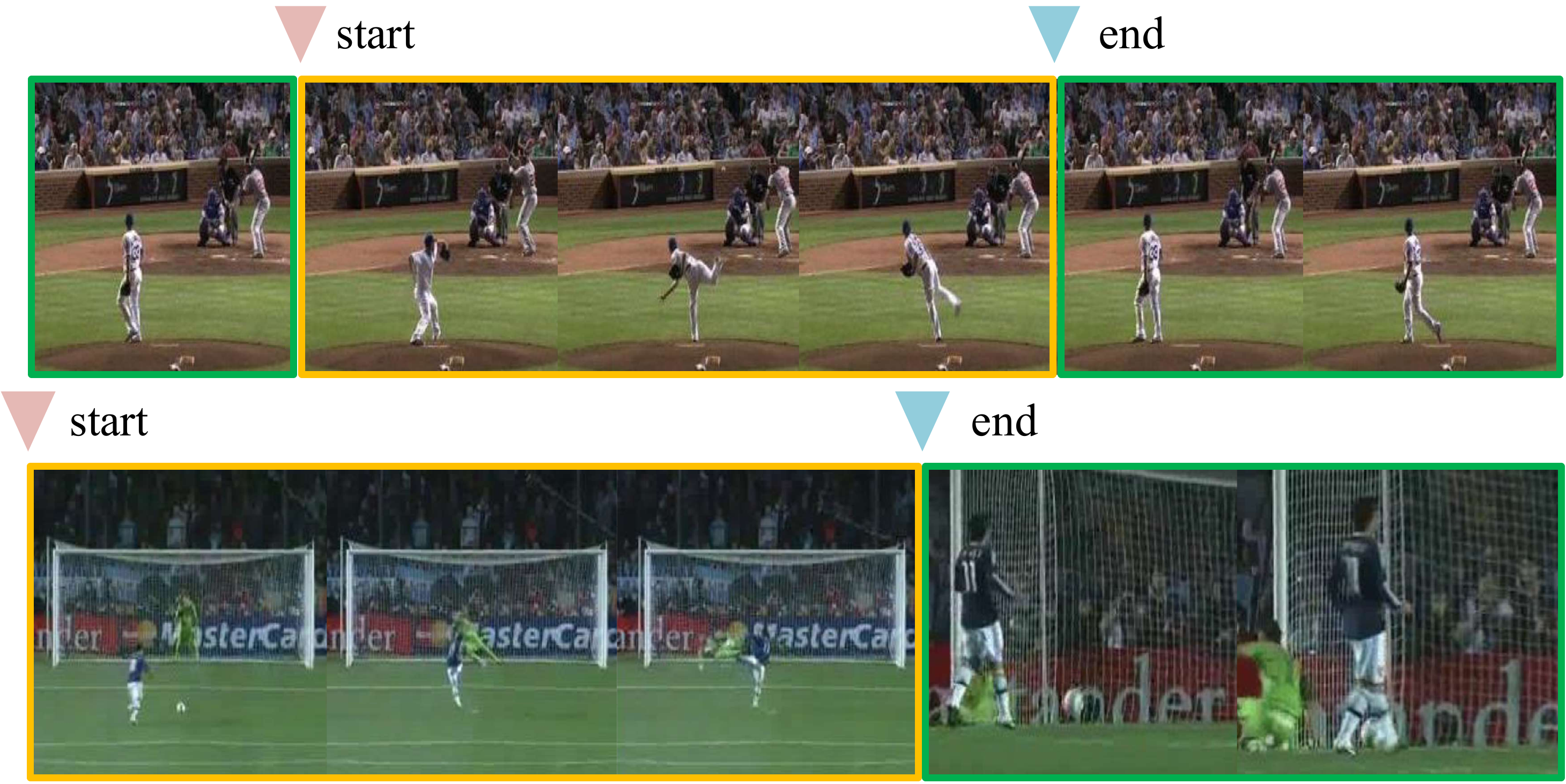}
	\caption{Examples of action samples~(yellow) and coupling samples~(green) for ``Baseball Pitch''~(top) and ``Soccer Penalty''~(bottom).}
	\label{fig:samples}
\end{figure}

For an untrimmed training video with its annotation $\Psi$, we can divide all snippets of the video into action snippets and non-action snippets, depending on whether they are within the extent of an action instance. 
%Subsequently, we collect every snippet within the start $t_{s,n}$ and end $t_{e,n}$ of an annotated action instance~$\varphi_{n}$ as an action sample, denote its feature vector as $\boldsymbol{a}$, and record its category label $c_n$. 
Subsequently, we treat action snippets of an action instance as an action sample, denote its feature vector as $\boldsymbol{a}$, and record its category label. We collect all action samples from the video.
To enable feature decoupling, we collect high-quality coupling samples through retrieving the non-action snippets that have high cosine similarities with at least one action sample. Accordingly, the feature vector of a coupling sample is denoted as $\boldsymbol{c}$. 
Finally, we obtain a collection of action samples $\mathbb{A}$ and a collection of coupling samples $\mathbb{C}$ from all training videos. 
% Note each coupling sample is paired with one (most similar) action sample, but not the other way round. 
We report some visual examples of action samples and coupling samples in Figure~\ref{fig:samples}.

\subsection{Decoupling Action and Co-occurrence Components}
We now decouple the action component and the co-occurrence component from the snippet features. % is to effectively separate these two orthogonal representations.
%We first give up the practice to annotate the co-occurrence components within coupling samples, which is expensive.
%In addition, it is also inappropriate to regard coupling samples as negative samples, as they often contain useful context clues for action classification.
A coupling sample does not include any actual action information as it is a non-action snippet. But, by construction, its features are similar to those of the corresponding action sample, indicating the coupling sample includes co-occurrence information of the human action, e.g., Figure~\ref{fig:samples}.
%We observe that coupling samples only include co-occurrence information of human actions, but do not include the actual action information. 
In other words, an action sample and its coupling sample share similar co-occurrence components but differ in whether the action component is present. Therefore, we propose to exploit both the similarity and difference between action samples and their coupling samples to factorize action contents from the dominating co-occurrence features. 

As illustrated in Figure~\ref{fig:pipeline}, we design two encoders $\boldsymbol{\phi}_{\mathrm{A}}$ and $\boldsymbol{\phi}_{\mathrm{C}}$ for feature decoupling.  The two encoders have the same network structure but do not share weights. They take an action sample $\boldsymbol{a} \in \mathbb{A}$ and its coupling sample $\boldsymbol{c} \in \mathbb{C}$ as input, and output an action feature vector and a co-occurrence feature vector for each sample, respectively.
%Then, we introduce two cosine similarity loss functions $\mathcal{L}_\mathrm{A}$ and $\mathcal{L}_\mathrm{C}$ to measure the similarity and difference between action feature and co-occurrence feature from action sample and coupling sample.
%Specifically, the similarity loss function $\mathcal{L}_{\mathrm{A}}$ and the difference loss function $\mathcal{L}_{\mathrm{C}}$ are defined as follows:
They are learned by minimizing two loss functions below:
\begin{align}
\mathcal{L}_{\mathrm{A}} &=\max \left\{0, \cos \left\langle\boldsymbol{\phi}_{\mathrm{A}}(\boldsymbol{a}), \boldsymbol{\phi}_{\mathrm{A}}(\boldsymbol{c})\right\rangle\right\},
\\
\mathcal{L}_{\mathrm{C}} &=\left(1-\cos \left\langle\boldsymbol{\phi}_{\mathrm{C}}(\boldsymbol{a}), \boldsymbol{\phi}_{\mathrm{C}}(\boldsymbol{c})\right\rangle\right), 
\end{align}
where $\boldsymbol{\phi}_{\mathrm{A}}$ extracts the action component of a sample, and $\boldsymbol{\phi}_{\mathrm{C}}$ extracts the co-occurrence component of a sample. 
%Encoders $E_\mathrm{A}$ and $E_\mathrm{C}$ are two different neural networks without weight sharing.
$\mathcal{L}_{\mathrm{A}}$ is used to minimize the similarity between the action component in an action sample, \textit{i.e.}, $\boldsymbol{\phi}_{\mathrm{A}}(\boldsymbol{a})$, and the action component in its coupling sample, \textit{i.e.}, $\boldsymbol{\phi}_{\mathrm{A}}(\boldsymbol{c})$. 
% [\textbf{Tang: is a and c from the same video? Is a the most similar action snippet w.r.t. c?}]
This is because an action sample contains the actual action content while its coupling sample does not.
$\mathcal{L}_{\mathrm{C}}$ is used to maximize the similarity between the co-occurrence component of an action sample, \textit{i.e.}, $\boldsymbol{\phi}_{\mathrm{C}}(\boldsymbol{a})$, and the co-occurrence component of its coupling sample, \textit{i.e.}, $\boldsymbol{\phi}_{\mathrm{C}}(\boldsymbol{c})$. This is because an action sample shares the similar co-occurrence component with its coupling sample. By minimizing these two losses over all training videos, the two encoders are learned to decouple the action component and the co-occurrence component of an arbitrary snippet.

% where $\boldsymbol{\phi}_{\mathrm{A}}(\boldsymbol{a})$ and $\boldsymbol{\phi}_{\mathrm{A}}(\boldsymbol{c})$ are  action feature vectors of the action sample and the coupling sample extracted from $E_\mathrm{A}$, respectively. Accordingly, $\boldsymbol{\phi}_{\mathrm{C}}(\boldsymbol{a})$ and $\boldsymbol{\phi}_{\mathrm{C}}(\boldsymbol{c})$ are action and co-occurrence feature vectors from the action sample and coupling sample extracted by $E_\mathrm{C}$, respectively. Encoders $E_\mathrm{A}$ and $E_\mathrm{C}$ are two different neural networks without weight sharing.
% The parameters of two encoders can be learned by minimizing these two objective functions.

\subsection{Recombining Action and Co-occurrence Components}
Till now, we have trained two encoders $\boldsymbol{\phi}_{\mathrm{A}}$ and $\boldsymbol{\phi}_{\mathrm{C}}$ to decouple the action component and the co-occurrence component of a snippet representation.
However, only using the action component for TAL is suboptimal 
% [\textbf{Tang: are there experimental results to support this claim?}] 
because the relevant contextual information included in the co-occurrence component often serves as useful clues for action classification, ~\textit{e.g.}, the tennis court of ``Tennis Swing''. This is also validated in our experiment. So it is necessary to effectively recombine them after explicit decoupling.
Given a snippet representation $\mathbf{f}$, we could obtain $\boldsymbol{\phi}_{\mathrm{A}}(\mathbf{f})$ and $ \boldsymbol{\phi}_{\mathrm{C}}(\mathbf{f})$ from the two encoders, which are then synthesized into a new feature representation, denoted as $\mathbf{f}^{\prime}$:
\begin{align}
\mathbf{f}^{\prime} &= \boldsymbol{\phi}_{\mathrm{A}}(\boldsymbol{\textbf{f}}) \oplus \boldsymbol{\phi}_{\mathrm{C}}(\textbf{f}),
%\\
%\boldsymbol{c}^{\prime} &= G_\mathrm{C}(\boldsymbol{\phi}_{\mathrm{A}}(\boldsymbol{c}), \boldsymbol{\phi}_{\mathrm{C}}(\boldsymbol{c})).
\end{align}
where $\oplus$ represents concatenation. Consequently, we can obtain the new video feature sequence $\mathbf{F}^{\prime}$ via RefactorNet.

In addition, we expect that the recombined feature representation contains salient action signals.
%, which leads to more effective action detection.
%To this end, we introduce two loss functions for $E_\mathrm{C}$ and $G$ to synthesize the desired feature representation.
To avoid overwhelming co-occurrence features from degenerating the performance of action detectors, we adopt a regularization based on the KL divergence to encourage the co-occurrence component obtained from $\boldsymbol{\phi}_{\mathrm{C}}$ to be close to the normal distribution $\mathcal{N}(0,1)$.  Inspired by~\cite{lu2019unsupervised}, we formulate the KL divergence loss as follows:
\begin{equation}
\mathcal{L}_{\mathrm{KL}}=\frac{1}{2} \sum_{i=1}^{D}\left(\mu_{i}^{2}+\sigma_{i}^{2}-\log \left(\sigma_{i}^{2}\right)-1\right),
\end{equation}
where $\mu$ and $\sigma$ represent the mean and standard deviation of a co-occurrence feature, and $D$ is the feature dimension.
The KL divergence loss regularizes co-occurrence features by limiting the distribution range.
As a result, the network will suppress overwhelming context or background information.
% and use more relevant co-occurrence features to synthesize the new representation
%On the other hand, since co-occurrence components often hold valuable auxiliary clues which are beneficial to action classification, we introduce a classification loss function to encourage $G$ to selectively contain informative context features:
%\begin{equation}
%L_{cls}=-\sum_{c=1}^{C} q_{c} \log p\left(c \mid \mathbf{w}_{c} \boldsymbol{a}^{\prime}\right),
%\end{equation}
%where $L_{cls}$ is implemented by a cross-entropy loss, and $\mathbf{w}_{c}$ is the parameters of the classifier. $q_{c}$ is the index label with $q_{c}=1$ when $c=c_n$ and $q_{c}=0$ otherwise. $p(c\mid\mathbf{w}_{c}\boldsymbol{a}^{\prime})$ is a softmax function. The generator $G$ 

Finally, the network will output the new snippet representation containing the salient action component, which leads to more effective video action detection.

\textbf{Summary.}
Our RefactorNet aims to obtain an appropriate snippet representation with both a salient action component and a suitable co-occurrence component for TAL.  We expect it to reduce the action ambiguity and also avoid over-relying on the co-occurrence component. This refactoring mechanism  distinguishes the proposed approach from previous approaches for TAL.

%------------------------------------------------------------------------
\subsection{Action Detection}
After feature refactoring, we need to locate action instances based on the new video representation.
We adopt a boundary-based proposal generator for candidate proposal generation. For each proposal, we extract its features from snippet features within it. Afterwards, we employ a multilayer perceptron~(MLP) as a localization head to refine the boundaries of each proposal, and two other MLPs as classification heads to predict the category and the completeness of the proposal. The product of the classification score and the completeness score is used as the confidence score of each proposal. Finally, we adopt Soft-NMS~(soft non-maximum suppression) to suppress redundant proposals with high overlaps and obtain detection results.

\begin{table*}[t]
\centering
\resizebox{1.0\linewidth}{!}{
\setlength{\tabcolsep}{0.9em}%
\begin{tabular}{l|c|cccccc|cccc}
\toprule %\hline
\multirow{2}{*}{Model} & \multirow{2}{*}{Conference} & \multicolumn{6}{c|}{THUMOS14~(\%)} & \multicolumn{4}{c}{ActivityNet v1.3~(\%)} \\ 
%\specialrule{0em}{1.0pt}{1.0pt}
\cline{3-12}
\specialrule{0em}{1.2pt}{1.2pt}
   & & 0.3   & 0.4   & 0.5   & 0.6   & 0.7   & Avg.  & 0.5 & 0.75 & 0.95 & Avg.      \\ 
\midrule % \hline
%TURN~\cite{gao2017turn}& ICCV & 44.1 & 34.9 & 25.6 & — & — & — & — & —  & — & —  \\
% SSN~\cite{zhao2017temporal}& ICCV'17 & 51.0 & 41.0 & 29.8 & — & — & — & 43.2 & 28.7 & 5.6 & 28.3      \\
% CDC~\cite{shou2017cdc}& CVPR'17 & 40.1 & 29.4 & 23.3 & 13.1 & 7.9 & 22.8 & 45.3 & 26.0 & 0.2 & 23.8    \\
%R-C3D~\cite{xu2017r} & 44.8 & 35.6 & 28.9 & — & — & — & 26.8 & — & — & — \\
%Turn-tap~\cite{gao2017turn}   &&&&&& &&&&\\ 
TAL-Net~\cite{chao2018rethinking}& CVPR'18 & 53.2 & 48.5 & 42.8 & 33.8 & 20.8 & 39.8 & 38.2 & 18.3 & 1.3 & 20.2    \\
BSN~\cite{lin2018bsn}& ECCV'18 & 53.5 & 45.0 & 36.9 & 28.4 & 20.0 & 36.8 & 46.5 & 30.0 & 8.0 & 30.0 \\
GTAN~\cite{long2019gaussian}& CVPR'19 & 57.8 & 47.2 & 38.8 & — & — & — & 52.6 & 34.1 & 8.9 & 34.3   \\
MGG~\cite{liu2019multi}& CVPR'19 & 53.9 & 46.8 & 37.4 & 29.5 & 21.3 & 37.8 & — & — & — & — \\
P-GCN~\cite{zeng2019graph}& ICCV'19 & 63.6 & 57.8 & 49.1 & — & — & — & 42.9 & 28.1 & 2.5 & 27.0 \\
BMN~\cite{lin2019bmn}& ICCV'19 & 56.0 & 47.4 & 38.8 & 29.7 & 20.5 & 38.5 & 50.1 & 34.8 & 8.3 & 33.9  \\
%A2Net~\cite{yang2020revisiting}& T-IP & 58.6 & 54.1 & 45.5 & 32.5 & 17.2 & 41.6 & 43.6 & 28.7 & 3.7 & 27.8 \\
%PBRNet~\cite{liu2020progressive}& AAAI'20 & 58.5 & 54.6 & 51.3 & 41.8 & 29.5 & 47.1 & 54.0 & 35.0 & 9.0 & 35.0  \\
% DBG~\cite{lin2020fast}& AAAI'20 & 57.8 & 49.4 & 42.8 & 33.8 & 21.7 & 41.1 & — & — & — & —  \\ 
G-TAD~\cite{xu2020g}& CVPR'20 & 54.5 & 47.6 & 40.2 & 30.8 & 23.4 & 39.3 & 50.4 & 34.6 & 9.0 & 34.1 \\
BC-GNN~\cite{bai2020boundary}& ECCV'20 & 57.1 & 49.1 & 40.4 & 31.2 & 23.1 & 40.2 & 50.6 & 34.8 & \textbf{9.4} & 34.3 \\
BU-MR~\cite{zhao2020bottom}& ECCV'20 & 53.9 & 50.7 & 45.4 & 38.0 & 28.5 & 43.3 & 43.5 & 33.9 & 9.2 & 30.1\\
%BSN++~\cite{su2021bsn++} & AAAI'21    & 59.9 & 49.5 & 41.3 & 31.9 & 22.8 & 41.1 & 51.3 & 35.7 & 8.3 & 34.9   \\ 
TCANet~\cite{qing2021temporal}& CVPR'21 & 60.6 & 53.2 & 44.6 & 36.8 & 26.7 & 44.4 & 54.3 & 39.1 & 8.4 & 37.6 \\
MUSES~\cite{liu2021multi}& CVPR'21 & 68.9 & 64.0 & 56.9 & 46.3 & 31.0 & 53.4 & 50.0 & 35.0 & 6.6 & 34.0 \\
AFSD~\cite{lin2021learning}& CVPR'21 & 67.3 & 62.4 & 55.5 & 43.7 & 31.1 & 52.0 & 52.4 & 35.3 & 6.5 & 34.4  \\
ContextLoc~\cite{zhu2021enriching}& ICCV'21 & 68.3 & 63.8 & 54.3 & 41.8 & 26.2 & 50.9 & 56.0 & 35.2 & 3.6 & 34.2 \\
RTD-Net~\cite{tan2021relaxed} & ICCV'21 & 68.3 & 62.3 & 51.9 & 38.8 & 23.7 & 49.0 & 47.2 & 30.7 & 8.6 & 30.8 \\
VSGN~\cite{zhao2021video} & ICCV'21 & 66.7 & 60.4 & 52.4 & 41.0 & 30.4 & 50.2 & 52.4 & 36.0 & 8.4 & 35.1 \\
\hline
\specialrule{0em}{0.8pt}{0.8pt} 
RefactorNet & CVPR'22 & \textbf{70.7} & \textbf{65.4} & \textbf{58.6} & \textbf{47.0} & \textbf{32.1} & \textbf{54.8} & \textbf{56.6} & \textbf{40.7} & 7.4 & \textbf{38.6} \\ 
\bottomrule 
\end{tabular}
}
\caption{Performance comparison on THUMOS14 and ActivityNet v1.3 in terms of mAP at different IoU thresholds. The ``Avg'' columns denote the average mAP in $[0.3:0.1:0.7]$ on THUMOS14 and $[0.5:0.05:0.95]$ on ActivityNet v1.3, respectively.}
\label{tab:comparison-on-thu-anet}
\end{table*}
%(*) represents the results of our implementation based on the code provided by the method.

\section{Training and Inference}
\subsection{Training}
During the training phase, we minimize the sum of the aforementioned losses for RefactorNet:
\begin{equation}
\mathcal{L}_{\mathcal{R}} = \alpha (\mathcal{L}_\mathrm{A} + \mathcal{L}_\mathrm{C}) +  \beta \mathcal{L}_\mathrm{KL},
\end{equation}
where $\alpha$ and $\beta$ are the weighting hyper-parameters.
%for each loss.

Next, we apply a boundary-based proposal generator,~\textit{e.g.}, BSN~\cite{lin2018bsn}, to the new video representation to produce a set of proposals. Especially, we define the following loss function for training the start and end action boundary predictors:
\begin{equation}
\mathcal{L}_{\mathcal{P}}= \mathcal{L}_{bl}^{\text {s}}+ \gamma \mathcal{L}_{bl}^{\text {e}},
\end{equation}
where $\gamma$ is a hyper-parameter. We adopt the binary logistic regression loss function as $\mathcal{L}_{bl}$. Then, we group probability peaks of starts and ends to produce candidate proposals. The features of each action proposal are extracted by RoI Pooling~\cite{girshick2015fast}.
In addition, the overall loss function for training the proposal refinement network is as follows: 
\begin{equation}
\mathcal{L}_{\mathcal{D}}=\mathcal{L}_{cls}+\lambda_{1} \mathcal{L}_{com}+\lambda_{2} \mathcal{L}_{reg},
\end{equation}
where $\lambda_{1}$ and $\lambda_{2}$ are hyper-parameters to trade-off these losses. 
We adopt the cross-entropy loss as $\mathcal{L}_{cls}$ for action classification, the hinge loss as $\mathcal{L}_{com}$ to predict the completeness score of each proposal, and the smooth-$\mathcal{L}_{1}$ loss as $\mathcal{L}_{reg}$ to predict the offset of center
coordinate and duration for each proposal. 

\noindent\subsection{Inference}
During the inference phase, given a feature vector $\mathbf{f}$ of a snippet in an untrimmed video, our RefactorNet takes it as input and outputs the new snippet feature vector $\mathbf{f}^{\prime}$ with refactored action and co-occurrence components. 
%by $G(\boldsymbol{\phi}_{\mathrm{A}}(\boldsymbol{a}), \boldsymbol{\phi}_{\mathrm{C}}(\boldsymbol{a}))$.
%Notably, the classification score represents the probability that the snippet contains an action.
Subsequently, we perform action boundary regression via the new snippet representations~$\textbf{F}^{\prime}$ to find start and end locations with high responses and group them into action proposals. The localization head and classification heads produce the offsets, category score and completeness score for each proposal, respectively.
We multiply the classification score and completeness score to get its confidence score. The final action localization result is retrieved by soft-NMS in the post-processing process.

\section{Experiments}
%We conduct extensive experiments and compare with previous works on THUMOS14 and ActivityNet v1.3 datasets. 
\subsection{Datasets and Metrics}
\noindent\textbf{THUMOS14}~\cite{jiang2014thumos}
is a standard benchmark for TAL. It contains 200 validation videos and 213 testing videos, including 20 action categories. It is very challenging since each video has more than 15 action instances. Following the common setting~\cite{zeng2021graph}, we use the validation set for training and evaluate on the testing set. 

\noindent\textbf{ActivityNet v1.3}~\cite{caba2015activitynet}
is a large-scale benchmark for video-based action localization. It contains 10k training videos and 5k validation videos corresponding to 200 different actions. Following the standard practice~\cite{liu2021multi}, we train our method on the training set and test it on the validation set.

\noindent\textbf{Evaluation Metrics.}
We use the mean Average Precision~(mAP) as the evaluation metric. The tIoU thresholds are $[0.3:0.1:0.7]$ for THUMOS14 and $[0.5:0.05:0.95]$ for ActivityNet v1.3. We also report the average mAP of the IoU thresholds between 0.5 and 0.95 with the step of 0.05 on ActivityNet v1.3.

\subsection{Implementation Details }
We divide each input video into 16-frame snippets and exploit a two-stream I3D network~\cite{carreira2017quo} pre-trained on the Kinetics dataset~\cite{kay2017kinetics} to extract original snippet features. 
%For RefactorNet, we empirically collect five coupling samples for each action sample during the training phase.
For RefactorNet, we adopt the Adam~\cite{kingma2014adam} optimizer to train the network. Similar to the training scheme~\cite{eom2019learning}, we train RefactorNet in two stages. In the first stage, we train the two encoders $\boldsymbol{\phi}_{\mathrm{A}}$ and $\boldsymbol{\phi}_{\mathrm{C}}$ with the corresponding losses $\mathcal{L}_\mathrm{A}$ and $\mathcal{L}_\mathrm{C}$ for 30 epochs on the training data with a learning rate 0.001. In the second stage, we train the whole network end to end with the learning rate of 0.001 for 20 epochs. 
We use the similarity score between a pair of an action sample and a coupling sample as $\alpha$ and empirically set $\beta=0.001$.
%In particular, we pay extra attention to $\mathcal{L}_\mathrm{A}$ and $\mathcal{L}_\mathrm{C}$, which are easy to overfit and separate unexpected feature vectors without appropriate intervention. 
%We find that it is difficult to adjust hyper-parameters for $\mathcal{L}_\mathrm{A}$ or $\mathcal{L}_\mathrm{C}$. Therefore, we use the similarity score between a pair of an action sample and a coupling sample as $\alpha$ and empirically set $\beta=0.001$. 
For the action detector, we adopt BSN~\cite{lin2018bsn} on THUMOS14 and ActivityNet v1.3 to produce action proposals. For each proposal, a feature representation is extracted via RoI pooling~\cite{ren2017faster}. Then, we apply three different MLPs on each proposal for boundary regression, action classification and completeness prediction, as in ~\cite{zeng2019graph, liu2021multi}. For $\mathcal{L}_\mathcal{D}$, we set $\gamma=1$ and $\lambda_{1}=\lambda_{2}=0.5$, as in~\cite{lin2018bsn, liu2021multi}. We train the model for 20 epochs with an initial learning rate 0.001.
For fair comparison, we combine our proposals with video-level classification results from~\cite{xiong2016cuhk} on ActivityNet v1.3.

%-------------------------------------------------------------------------

\subsection{Comparison with State-of-the-art Methods}
\noindent\textbf{THUMOS14.} Our method is compared with state-of-the-art methods in Table~\ref{tab:comparison-on-thu-anet}. 
We report mAP at different tIoU thresholds as well as average mAP between 0.3 and 0.7 with the step of 0.1.
Especially, the mAP of our method is 1.7\% higher than that of MUSES~\cite{liu2021multi} when tIoU=0.5.
Our method also achieves significant improvement on the average mAP. It indicates the effectiveness of our method for accurate action localization. 

\noindent\textbf{ActivityNet v1.3.} We also report the action localization results of various methods in Table~\ref{tab:comparison-on-thu-anet}. At tIoU 0.5, our method outperforms all the other methods. In addition, our improvement is particularly notable at tIoU=0.75, where we outperform the state-of-the-art methods by a margin of 1.6\%. These experimental results further demonstrate the effectiveness of our method.
%-------------------------------------------------------------------------

\subsection{Ablation Study}

In this section, we conduct comprehensive ablation studies in two folds. On the one hand, we verify the effects of main components within the proposed model.
On the other hand, we verify the effectiveness of our method for temporal action localization and report both qualitative and quantitative experimental results and analyses.

%-------------------------------------------------------------------------
\begin{table}[t]
\centering
\resizebox{1.0\linewidth}{!}{
\setlength{\tabcolsep}{0.5em}%
\begin{tabular}{l|cccccc}
\toprule %\hline
\multirow{2}{*}{Method} & \multicolumn{6}{c}{mAP@tIoU (\%)} \\
& 0.3  & 0.4 & 0.5 & 0.6 & 0.7 & Avg. \\ 
\midrule % \hline
Baseline & 68.2 & 62.7 & \underline{56.0} & \underline{45.9} & \underline{30.1} & \underline{52.6} \\
Deep Baseline & \underline{68.5} & \underline{63.0} & 55.5 & 45.3 & 29.2 & 52.3    \\
Baseline + RefactorNet & \textbf{70.7} & \textbf{65.4} & \textbf{58.6} & \textbf{47.0} & \textbf{32.1} & \textbf{54.8}    \\
\bottomrule 
\end{tabular}
}
\caption{
Ablation study of the effect of feature refactoring on THUMOS14. The baseline is constructed by removing RefactorNet from our framework. The deep baseline is constructed by replacing RefactorNet in our framework with temporal 1D convolutional layers so that its model size and depth are the same as those of our framework.
%Ablation studies of fair comparison with the localization baseline on THUMOS14. ``Deep Baseline'' indicates the result of introducing the same number of parameters as RefactorNet for the baseline.
}
\label{tab:ablation-on-parameter}
\end{table}
\noindent\textbf{Effect of Feature Refactoring.} 
We construct a baseline by removing feature refactoring from our model. 
Table~\ref{tab:ablation-on-parameter} shows our model outperforms the baseline by a large margin.
For fair comparison, we further construct a deep baseline by adding more layers (\emph{i.e.}, temporal 1D convolutional layers) to the baseline so that its model size and depth are the same as those of our model.
%The total number of learnable parameters in our RefactorNet is 15.8M. To make a fair comparison with the localization baseline without RefactorNet, we introduce the same number of learnable parameters into the baseline and report comparison results in Table~\ref{tab:ablation-on-parameter}. 
Table~\ref{tab:ablation-on-parameter} shows naively increasing the model size and network depth does not improve the baseline. This indicates that our performance improvement comes from effectively refactoring feature representations rather than simply adding more layers.

%-------------------------------------------------------------------------
\begin{table}[t]
\centering
\resizebox{0.8\linewidth}{!}{
\setlength{\tabcolsep}{0.8em}%
\begin{tabular}{l|c|c}
\toprule %\hline
Method & Class (\%) & tIoU (\%) \\
%\\ & 0.1  & 0.2 & 0.3 & 0.4 & 0.5 & Avg. \\ 
\midrule % \hline
Baseline & 76.7 & 79.6  \\
Deep Baseline & 77.1 & 78.4 \\
Baseline + RefactorNet & \textbf{85.9} & \textbf{83.5}  \\
\bottomrule 
\end{tabular}
}
\caption{Ablation study of action classification accuracy (Class) and localization accuracy (tIoU) of high-quality proposals, whose tIoU w.r.t. ground truth are greater than 0.7, on THUMOS14.}
\label{tab:ablation-on-performance}
\end{table}
\noindent\textbf{Analysis of Performance Improvements.}
In order to further explore the reasons behind performance improvements, we report the accuracy of action classification and the average tIoU of high-quality proposals in Table~\ref{tab:ablation-on-performance}. High-quality proposals refer to action proposals after the regression whose tIoU w.r.t. ground truth is greater than 0.7. The experimental results show that the refactored snippet representations benefit both action classification and temporal boundary regression, compared with the baseline and deep baseline.

%-------------------------------------------------------------------------
\begin{table}[t]
\centering
\resizebox{1.0\linewidth}{!}{
\setlength{\tabcolsep}{0.75em}%
\begin{tabular}{cc|cccccc}
\toprule %\hline
\specialrule{0em}{1.0pt}{1.0pt}
\multirow{2}{*}{$\mathcal{L}_{\mathrm{A}} \& \mathcal{L}_{\mathrm{C}}$} & \multirow{2}{*}{$\mathcal{L}_\mathrm{KL}$} & \multicolumn{6}{c}{mAP@tIoU (\%)}
\\& & 0.3  & 0.4 & 0.5 & 0.6 & 0.7 & Avg. \\ 
\midrule % \hline
  &  & 68.2 & 62.7 & 56.0 & 45.9 & 30.1 & 52.6      \\
\Checkmark   &  & 69.6 & 64.2 & 57.8 & 46.1 & 31.4 & 53.8    \\
\Checkmark   & \Checkmark  & \textbf{70.7} & \textbf{65.4} & \textbf{58.6} & \textbf{47.0} & \textbf{32.1} & \textbf{54.8}    \\
\bottomrule 
\end{tabular}
}
\caption{Ablation study of different losses on THUMOS14. $\mathcal{L}_{\mathrm{A}}$ and $\mathcal{L}_{\mathrm{C}}$ are used to decouple action and co-occurrence components and $\mathcal{L}_\mathrm{KL}$ is used to regularize the co-occurrence component.}
\label{tab:ablation-on-loss}
\end{table}
% in terms of mAP~(\%)
\noindent\textbf{Loss Functions.}
To verify the effectiveness of our different loss functions, we conduct an ablation experiment by using different losses to train our model. In Table~\ref{tab:ablation-on-loss}, we measure mAP and report results on THUMOS14. The first row demonstrates the performance of our baseline, which does not include our RefactorNet.
The second row indicates that RefactorNet is guided by $\mathcal{L}_\mathrm{A}$ and $\mathcal{L}_\mathrm{C}$ to explicitly decouple the actual action from its co-occurrence components.
The experimental results show that explicit feature refactoring can significantly facilitate action localization. In addition, $\mathcal{L}_\mathrm{KL}$ further regularizes the co-occurrence features by limiting their distribution range.
The third row in Table~\ref{tab:ablation-on-loss} demonstrates that $\mathcal{L}_\mathrm{KL}$ will enforce the network to use more action information when synthesizing the new video representation for action localization. 
With the combination of these loss functions, our RefactorNet is driven to combine the action component and the co-occurrence component within each snippet in a more balanced way. As a result, the action detector can achieve remarkable performance.

\begin{figure}[t]
	\centering
	\includegraphics[width=\linewidth]{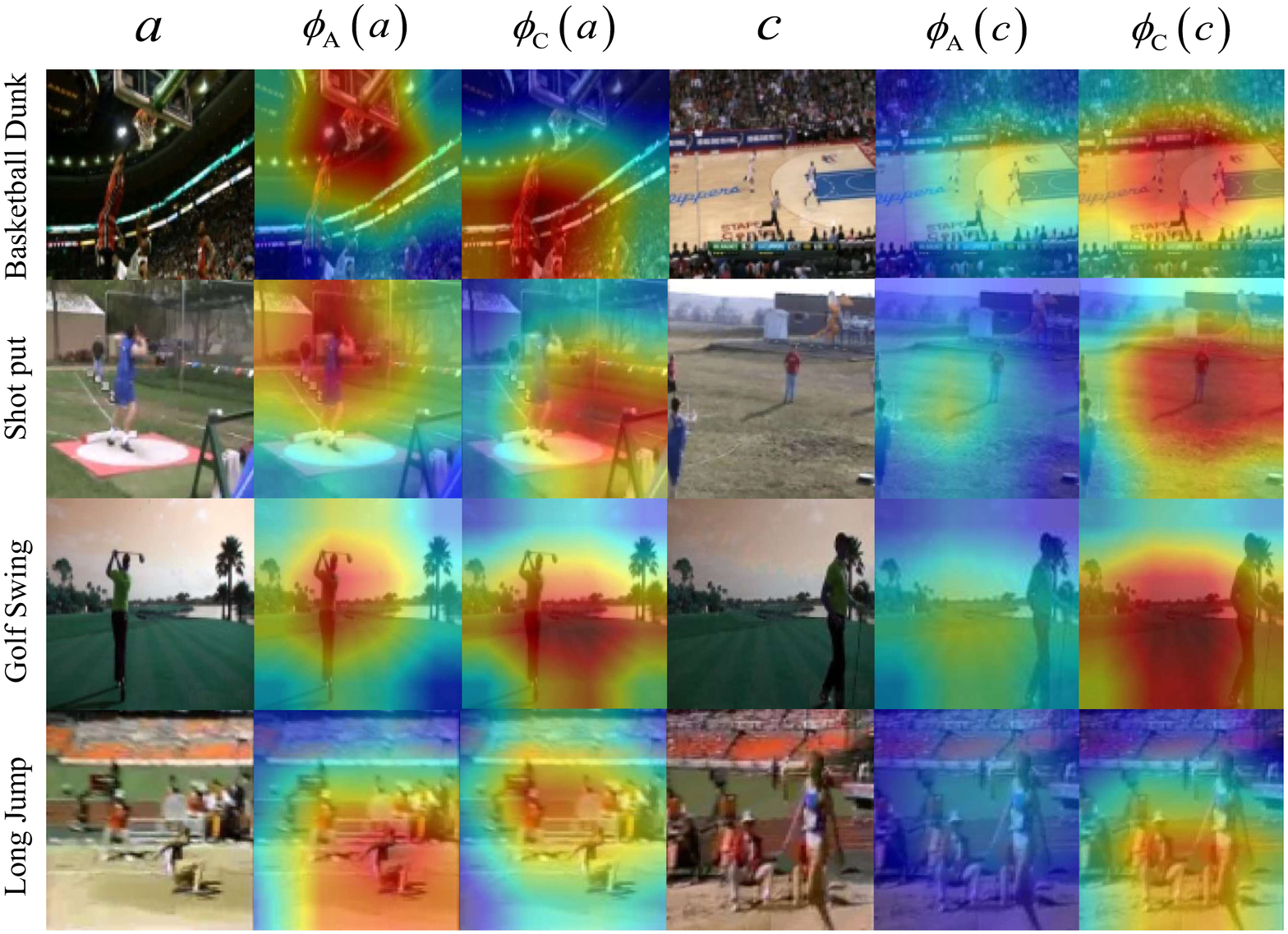}
	\caption{Visualization of the action content and the corresponding co-occurring component decoupled by our RefactorNet from action samples and coupling samples. The decoupled action and co-occurrence components respectively focus on the action and co-occurring visual elements.}
	\label{fig:vis}
\end{figure}

%-------------------------------------------------------------------------
\noindent\textbf{Visual Analysis for Decoupled Features.} To validate the ability of our method to decouple the action component and the co-occurrence component, we conduct spatial visualization for the outputs of two encoders, \textit{i.e.}, $\boldsymbol{\phi}_{\mathrm{A}}(\boldsymbol{a})$, $\boldsymbol{\phi}_{\mathrm{C}}(\boldsymbol{a})$,  $\boldsymbol{\phi}_{\mathrm{A}}(\boldsymbol{c})$ and $\boldsymbol{\phi}_{\mathrm{C}}(\boldsymbol{c})$, by class activation maps~(CAM~\cite{zhou2016learning}).
As illustrated in Figure~\ref{fig:vis}, the outputs of two encoders are visualized by heatmaps imposed on the original frames.
It can be observed that the action representation obtained by the encoder $\boldsymbol{\phi}_{\mathrm{A}}$ focuses on the action area,~\textit{e.g.}, dunk of ``Basketball Dunk'' or swing of ``Golf Swing'', and the co-occurrence representation obtained by the encoder $\boldsymbol{\phi}_{\mathrm{C}}$ focuses on the background scene, \textit{e.g.}, lawn or sand. 
As a result, our method can effectively decouple action and co-occurrence components, and recombine them to facilitate action localization.

\begin{figure}[t]
	\centering
	\includegraphics[width=\linewidth]{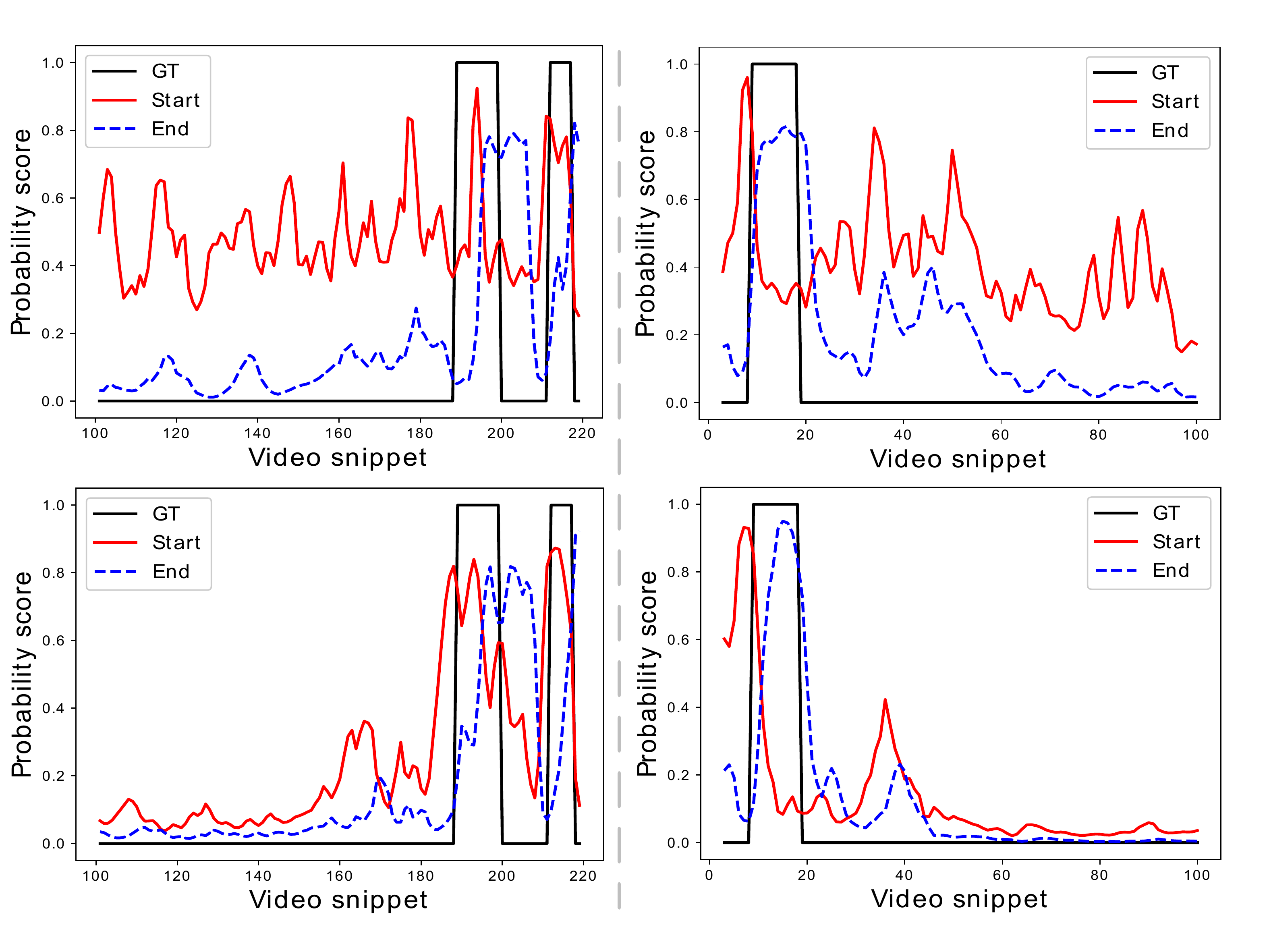}
	\caption{Comparison of temporal boundary predictions produced by the baseline~(top) and our method~(bottom). Particularly, the average precision of action boundary prediction of our method is 19.3\%, while 13.4\% for the baseline. It can be observed that our RefactorNet has better robustness and higher precision.}
	\label{fig:prediction}
\end{figure}

%-------------------------------------------------------------------------
\noindent\textbf{Visual Analysis for Temporal Boundary Prediction.}
Co-occurrence components often dominate action components in video snippets. This causes unreliable boundary predictions that degenerate detection performance.
In order to verify whether our method can facilitate robust boundary prediction, we visualize some prediction results in Figure~\ref{fig:prediction}. Compared with the baseline, our method can help accurately predict boundary locations as well as reduce false alarms caused by the distraction of co-occurrence components.
Visualization examples demonstrate that regulating co-occurrence components is an effective solution for robust temporal boundary prediction.

\begin{figure}[t]
	\centering
	\includegraphics[width=\linewidth]{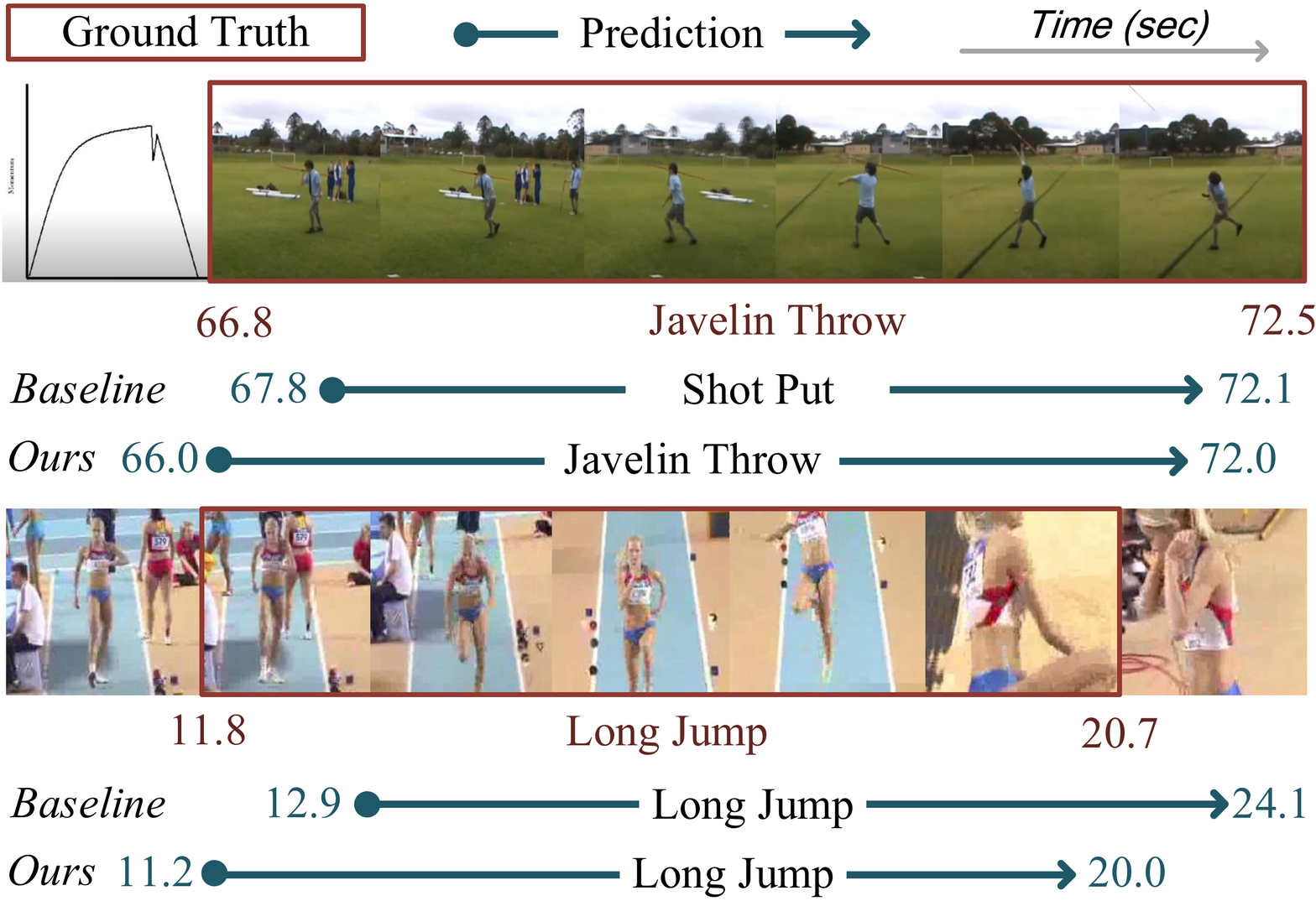}
	\caption{Qualitative detection results on ActivityNet v1.3 (top) and THUMOS14 (bottom).}
	\label{fig:quanlitative}
\end{figure}

\noindent\textbf{Visual Analysis for Temporal Action Localization. }
To verify whether the new video representation contains more favorable clues for action localization, we report some qualitative results in Figure~\ref{fig:quanlitative}. We can see that our method can help the model effectively learn indicative information to alleviate the misclassification caused by over-relying on the co-occurrence component. In addition, refactoring action and co-occurrence components can provide salient action features to reduce the action ambiguity and uncertainty.
%-------------------------------------------------------------------------

%-------------------------------------------------------------------------
\begin{table}[t]
\centering
\resizebox{0.8\linewidth}{!}{
\setlength{\tabcolsep}{0.65em}%
\begin{tabular}{l|ccc|c}
\toprule %\hline
\specialrule{0em}{1.0pt}{1.0pt}
\multirow{2}{*}{Modality} & \multicolumn{3}{c|}{mAP@tIoU (\%)} & \multirow{2}{*}{Class (\%)}
\\ 
& 0.3  & 0.5  & 0.7  \\ 
\specialrule{0em}{.5pt}{.5pt}
\midrule % \hline
RGB \& Flow  & 68.2 & 56.0  & 30.1 & 76.7     \\
Only Action     & 69.6  & 57.2  & 30.9 & 81.3   \\
Action \& Co-occ.   & \textbf{70.7} & \textbf{58.6}  & \textbf{32.1} & \textbf{85.9}    \\
\bottomrule 
\end{tabular}
}
\caption{Ablation study of  using only the action component for TAL in terms of mAP at different IoU thresholds and the classification accuracy (Class) on THUMOS14. ``RGB \& Flow'' represents the original video features. ``Action'' and ``Co-occ.'' represent the action component and the co-occurrence component, respectively.}
\label{tab:ablation-on-component}
\end{table}
% in terms of mAP~(\%)
\noindent\textbf{Analysis of the Action Component.}
TAL not only needs to locate actions but also accurately predict their categories.
Table~\ref{tab:ablation-on-component} indicates that only using the action component for TAL is suboptimal because the co-occurrence component often contains contextual information useful to action classification, and helps reduce action ambiguity and uncertainty. 
%Meanwhile, co-occurrence components can also help reduce action ambiguity and uncertainty.
%\noindent\textbf{Failure Analysis.}
%-------------------------------------------------------------------------
\section{Conclusion}

In this paper, we rethink and explore refactoring action and co-occurrence components of snippet features for TAL. We present a novel feature refactoring network, RefactorNet, inserted between the video feature extractor and the action detector. It aims at decoupling the snippet representation into the actual action content and its co-occurrence component, which are then recombined into a new snippet representation with salient action component and suitable co-occurrence component.
%the actual action content from its co-occurrence components, and refactoring new video representation with salient action component and suitable co-occurrence component. 
Quantitative and qualitative experiments verify the effectiveness of the proposed method.
As a result, the combination of RefactorNet with action detectors achieves great performance on THUMOS14 and ActivityNet v1.3.
%-------------------------------------------------------------------------

\section{Acknowledgment}
This work was supported partly by National Key R\&D Program of China under Grant 2018AAA0101400, NSFC under Grants 62088102, 61976171, and 62106192, Natural Science Foundation of Shaanxi Province under Grants 2022JC-41 and 2021JQ-054, China Postdoctoral Science Foundation under Grant 2020M683490, and Fundamental Research Funds for the Central Universities under Grant XTR042021005.

%%%%%%%%% REFERENCES
{\small
\bibliographystyle{ieee_fullname}
\bibliography{cvpr22}

\begin{thebibliography}{10}\itemsep=-1pt

\bibitem{bahng2020learning}
Hyojin Bahng, Sanghyuk Chun, Sangdoo Yun, Jaegul Choo, and Seong~Joon Oh.
\newblock Learning de-biased representations with biased representations.
\newblock In {\em ICML}, pages 528--539, 2020.

\bibitem{bai2020boundary}
Yueran Bai, Yingying Wang, Yunhai Tong, Yang Yang, Qiyue Liu, and Junhui Liu.
\newblock Boundary content graph neural network for temporal action proposal
  generation.
\newblock In {\em ECCV}, pages 121--137, 2020.

\bibitem{caba2015activitynet}
Fabian Caba~Heilbron, Victor Escorcia, Bernard Ghanem, and Juan Carlos~Niebles.
\newblock Activitynet: A large-scale video benchmark for human activity
  understanding.
\newblock In {\em CVPR}, pages 961--970, 2015.

\bibitem{carreira2017quo}
Joao Carreira and Andrew Zisserman.
\newblock Quo vadis, action recognition? a new model and the kinetics dataset.
\newblock In {\em CVPR}, pages 6299--6308, 2017.

\bibitem{chao2018rethinking}
Yu-Wei Chao, Sudheendra Vijayanarasimhan, Bryan Seybold, David~A Ross, Jia
  Deng, and Rahul Sukthankar.
\newblock Rethinking the faster r-cnn architecture for temporal action
  localization.
\newblock In {\em CVPR}, pages 1130--1139, 2018.

\bibitem{choi2019can}
Jinwoo Choi, Chen Gao, Joseph~CE Messou, and Jia-Bin Huang.
\newblock Why can't i dance in a mall? learning to mitigate scene bias in
  action recognition.
\newblock In {\em NeurIPS}, pages 853--865, 2019.

\bibitem{denton2017unsupervised}
Emily~L Denton and Vighnesh Birodkar.
\newblock Unsupervised learning of disentangled representations from video.
\newblock In {\em NIPS}, pages 4417--4426, 2017.

\bibitem{eom2019learning}
Chanho Eom and Bumsub Ham.
\newblock Learning disentangled representation for robust person
  re-identification.
\newblock In {\em NeurIPS}, pages 5297--5308, 2019.

\bibitem{fan2018end}
Lijie Fan, Wenbing Huang, Chuang Gan, Stefano Ermon, Boqing Gong, and Junzhou
  Huang.
\newblock End-to-end learning of motion representation for video understanding.
\newblock In {\em CVPR}, pages 6016--6025, 2018.

\bibitem{gao2020accurate}
Jialin Gao, Zhixiang Shi, Guanshuo Wang, Jiani Li, Yufeng Yuan, Shiming Ge, and
  Xi Zhou.
\newblock Accurate temporal action proposal generation with relation-aware
  pyramid network.
\newblock In {\em AAAI}, pages 10810--10817, 2020.

\bibitem{girshick2015fast}
Ross Girshick.
\newblock Fast r-cnn.
\newblock In {\em ICCV}, pages 1440--1448, 2015.

\bibitem{gong2020scale}
Guoqiang Gong, Liangfeng Zheng, and Yadong Mu.
\newblock Scale matters: Temporal scale aggregation network for precise action
  localization in untrimmed videos.
\newblock In {\em ICME}, pages 1--6, 2020.

\bibitem{hamaguchi2019rare}
Ryuhei Hamaguchi, Ken Sakurada, and Ryosuke Nakamura.
\newblock Rare event detection using disentangled representation learning.
\newblock In {\em CVPR}, pages 9327--9335, 2019.

\bibitem{hsieh2018learning}
Jun-Ting Hsieh, Bingbin Liu, De-An Huang, Fei-Fei Li, and Juan~Carlos Niebles.
\newblock Learning to decompose and disentangle representations for video
  prediction.
\newblock In {\em NeurIPS}, pages 515--524, 2018.

\bibitem{huang2021self}
Lianghua Huang, Yu Liu, Bin Wang, Pan Pan, Yinghui Xu, and Rong Jin.
\newblock Self-supervised video representation learning by context and motion
  decoupling.
\newblock In {\em CVPR}, pages 13886--13895, 2021.

\bibitem{jain2020actionbytes}
Mihir Jain, Amir Ghodrati, and Cees~GM Snoek.
\newblock Actionbytes: Learning from trimmed videos to localize actions.
\newblock In {\em CVPR}, pages 1171--1180, 2020.

\bibitem{jiang2014thumos}
Yu-Gang Jiang, Jingen Liu, A~Roshan Zamir, George Toderici, Ivan Laptev,
  Mubarak Shah, and Rahul Sukthankar.
\newblock Thumos challenge: Action recognition with a large number of classes,
  2014.

\bibitem{kay2017kinetics}
Will Kay, Joao Carreira, Karen Simonyan, Brian Zhang, Chloe Hillier, Sudheendra
  Vijayanarasimhan, Fabio Viola, Tim Green, Trevor Back, Paul Natsev, et~al.
\newblock The kinetics human action video dataset.
\newblock {\em arXiv preprint arXiv:1705.06950}, 2017.

\bibitem{kingma2014adam}
Diederik~P Kingma and Jimmy Ba.
\newblock Adam: A method for stochastic optimization.
\newblock In {\em ICLR}, 2015.

\bibitem{lee2020background}
Pilhyeon Lee, Youngjung Uh, and Hyeran Byun.
\newblock Background suppression network for weakly-supervised temporal action
  localization.
\newblock In {\em AAAI}, pages 11320--11327, 2020.

\bibitem{lin2020fast}
Chuming Lin, Jian Li, Yabiao Wang, Ying Tai, Donghao Luo, Zhipeng Cui, Chengjie
  Wang, Jilin Li, Feiyue Huang, and Rongrong Ji.
\newblock Fast learning of temporal action proposal via dense boundary
  generator.
\newblock In {\em AAAI}, pages 11499--11506, 2020.

\bibitem{lin2021learning}
Chuming Lin, Chengming Xu, Donghao Luo, Yabiao Wang, Ying Tai, Chengjie Wang,
  Jilin Li, Feiyue Huang, and Yanwei Fu.
\newblock Learning salient boundary feature for anchor-free temporal action
  localization.
\newblock In {\em CVPR}, pages 3320--3329, 2021.

\bibitem{lin2019bmn}
Tianwei Lin, Xiao Liu, Xin Li, Errui Ding, and Shilei Wen.
\newblock Bmn: Boundary-matching network for temporal action proposal
  generation.
\newblock In {\em ICCV}, pages 3889--3898, 2019.

\bibitem{lin2018bsn}
Tianwei Lin, Xu Zhao, Haisheng Su, Chongjing Wang, and Ming Yang.
\newblock Bsn: Boundary sensitive network for temporal action proposal
  generation.
\newblock In {\em ECCV}, pages 3--19, 2018.

\bibitem{liu2021unsupervised}
Shilong Liu, Lei Zhang, Xiao Yang, Hang Su, and Jun Zhu.
\newblock Unsupervised part segmentation through disentangling appearance and
  shape.
\newblock In {\em CVPR}, pages 8355--8364, 2021.

\bibitem{liu2021multi}
Xiaolong Liu, Yao Hu, Song Bai, Fei Ding, Xiang Bai, and Philip~HS Torr.
\newblock Multi-shot temporal event localization: a benchmark.
\newblock In {\em CVPR}, pages 12596--12606, 2021.

\bibitem{liu2019multi}
Yuan Liu, Lin Ma, Yifeng Zhang, Wei Liu, and Shih-Fu Chang.
\newblock Multi-granularity generator for temporal action proposal.
\newblock In {\em CVPR}, pages 3604--3613, 2019.

\bibitem{liu2018exploring}
Yu Liu, Fangyin Wei, Jing Shao, Lu Sheng, Junjie Yan, and Xiaogang Wang.
\newblock Exploring disentangled feature representation beyond face
  identification.
\newblock In {\em CVPR}, pages 2080--2089, 2018.

\bibitem{liu2021weakly}
Ziyi Liu, Le Wang, Wei Tang, Junsong Yuan, Nanning Zheng, and Gang Hua.
\newblock Weakly supervised temporal action localization through learning
  explicit subspaces for action and context.
\newblock In {\em AAAI}, pages 2242--2250, 2021.

\bibitem{long2019gaussian}
Fuchen Long, Ting Yao, Zhaofan Qiu, Xinmei Tian, Jiebo Luo, and Tao Mei.
\newblock Gaussian temporal awareness networks for action localization.
\newblock In {\em CVPR}, pages 344--353, 2019.

\bibitem{lu2019unsupervised}
Boyu Lu, Jun-Cheng Chen, and Rama Chellappa.
\newblock Unsupervised domain-specific deblurring via disentangled
  representations.
\newblock In {\em CVPR}, pages 10225--10234, 2019.

\bibitem{pan2021videomoco}
Tian Pan, Yibing Song, Tianyu Yang, Wenhao Jiang, and Wei Liu.
\newblock Videomoco: Contrastive video representation learning with temporally
  adversarial examples.
\newblock In {\em CVPR}, pages 11205--11214, 2021.

\bibitem{qing2021temporal}
Zhiwu Qing, Haisheng Su, Weihao Gan, Dongliang Wang, Wei Wu, Xiang Wang, Yu
  Qiao, Junjie Yan, Changxin Gao, and Nong Sang.
\newblock Temporal context aggregation network for temporal action proposal
  refinement.
\newblock In {\em CVPR}, pages 485--494, 2021.

\bibitem{ren2017faster}
Shaoqing Ren, Kaiming He, Ross Girshick, and Jian Sun.
\newblock Faster r-cnn: Towards real-time object detection with region proposal
  networks.
\newblock {\em T-PAMI}, pages 1137--1149, 2017.

\bibitem{shou2017cdc}
Zheng Shou, Jonathan Chan, Alireza Zareian, Kazuyuki Miyazawa, and Shih-Fu
  Chang.
\newblock Cdc: Convolutional-de-convolutional networks for precise temporal
  action localization in untrimmed videos.
\newblock In {\em CVPR}, pages 5734--5743, 2017.

\bibitem{simonyan2014two}
Karen Simonyan and Andrew Zisserman.
\newblock Two-stream convolutional networks for action recognition in videos.
\newblock In {\em NeurIPS}, pages 568--576, 2014.

\bibitem{singh2020don}
Krishna~Kumar Singh, Dhruv Mahajan, Kristen Grauman, Yong~Jae Lee, Matt
  Feiszli, and Deepti Ghadiyaram.
\newblock Don't judge an object by its context: Learning to overcome contextual
  bias.
\newblock In {\em CVPR}, pages 11070--11078, 2020.

\bibitem{tan2021relaxed}
Jing Tan, Jiaqi Tang, Limin Wang, and Gangshan Wu.
\newblock Relaxed transformer decoders for direct action proposal generation.
\newblock {\em ICCV}, 2021.

\bibitem{tirupattur2021modeling}
Praveen Tirupattur, Kevin Duarte, Yogesh~S Rawat, and Mubarak Shah.
\newblock Modeling multi-label action dependencies for temporal action
  localization.
\newblock In {\em CVPR}, pages 1460--1470, 2021.

\bibitem{villegas2017decomposing}
Ruben Villegas, Jimei Yang, Seunghoon Hong, Xunyu Lin, and Honglak Lee.
\newblock Decomposing motion and content for natural video sequence prediction.
\newblock {\em ICLR}, 2017.

\bibitem{wang2021enhancing}
Jinpeng Wang, Yuting Gao, Ke Li, Xinyang Jiang, Xiaowei Guo, Rongrong Ji, and
  Xing Sun.
\newblock Enhancing unsupervised video representation learning by decoupling
  the scene and the motion.
\newblock In {\em AAAI}, pages 10129--10137, 2021.

\bibitem{wang2018pulling}
Yang Wang and Minh Hoai.
\newblock Pulling actions out of context: Explicit separation for effective
  combination.
\newblock In {\em CVPR}, pages 7044--7053, 2018.

\bibitem{wray2021semantic}
Michael Wray, Hazel Doughty, and Dima Damen.
\newblock On semantic similarity in video retrieval.
\newblock In {\em CVPR}, pages 3650--3660, 2021.

\bibitem{wu2019disentangled}
Xiang Wu, Huaibo Huang, Vishal~M Patel, Ran He, and Zhenan Sun.
\newblock Disentangled variational representation for heterogeneous face
  recognition.
\newblock In {\em AAAI}, pages 9005--9012, 2019.

\bibitem{xiong2016cuhk}
Yuanjun Xiong, Limin Wang, Zhe Wang, Bowen Zhang, Hang Song, Wei Li, Dahua Lin,
  Yu Qiao, Luc Van~Gool, and Xiaoou Tang.
\newblock Cuhk \& ethz \& siat submission to activitynet challenge 2016.
\newblock {\em arXiv preprint arXiv:1608.00797}, 2016.

\bibitem{xu2020g}
Mengmeng Xu, Chen Zhao, David~S Rojas, Ali Thabet, and Bernard Ghanem.
\newblock G-tad: Sub-graph localization for temporal action detection.
\newblock In {\em CVPR}, pages 10156--10165, 2020.

\bibitem{yang2018exploring}
Ke Yang, Peng Qiao, Dongsheng Li, Shaohe Lv, and Yong Dou.
\newblock Exploring temporal preservation networks for precise temporal action
  localization.
\newblock In {\em AAAI}, 2018.

\bibitem{yang2021background}
Le Yang, Junwei Han, Tao Zhao, Tianwei Lin, Dingwen Zhang, and Jianxin Chen.
\newblock Background-click supervision for temporal action localization.
\newblock {\em T-PAMI}, 2021.

\bibitem{yingzhen2018disentangled}
Li Yingzhen and Stephan Mandt.
\newblock Disentangled sequential autoencoder.
\newblock In {\em ICML}, pages 5670--5679, 2018.

\bibitem{yuan2017temporal}
Zehuan Yuan, Jonathan~C Stroud, Tong Lu, and Jia Deng.
\newblock Temporal action localization by structured maximal sums.
\newblock In {\em CVPR}, pages 3684--3692, 2017.

\bibitem{zeng2019graph}
Runhao Zeng, Wenbing Huang, Mingkui Tan, Yu Rong, Peilin Zhao, Junzhou Huang,
  and Chuang Gan.
\newblock Graph convolutional networks for temporal action localization.
\newblock In {\em ICCV}, pages 7094--7103, 2019.

\bibitem{zeng2021graph}
Runhao Zeng, Wenbing Huang, Mingkui Tan, Yu Rong, Peilin Zhao, Junzhou Huang,
  and Chuang Gan.
\newblock Graph convolutional module for temporal action localization in
  videos.
\newblock {\em IEEE Transactions on Pattern Analysis and Machine
  Intelligence~(T-PAMI)}, 2021.

\bibitem{zhai2020two}
Yuanhao Zhai, Le Wang, Wei Tang, Qilin Zhang, Junsong Yuan, and Gang Hua.
\newblock Two-stream consensus network for weakly-supervised temporal action
  localization.
\newblock In {\em ECCV}, 2020.

\bibitem{zhao2021video}
Chen Zhao, Ali~K Thabet, and Bernard Ghanem.
\newblock Video self-stitching graph network for temporal action localization.
\newblock In {\em ICCV}, pages 13658--13667, 2021.

\bibitem{zhao2020bottom}
Peisen Zhao, Lingxi Xie, Chen Ju, Ya Zhang, Yanfeng Wang, and Qi Tian.
\newblock Bottom-up temporal action localization with mutual regularization.
\newblock In {\em ECCV}, pages 539--555, 2020.

\bibitem{zhou2016learning}
Bolei Zhou, Aditya Khosla, Agata Lapedriza, Aude Oliva, and Antonio Torralba.
\newblock Learning deep features for discriminative localization.
\newblock In {\em CVPR}, pages 2921--2929, 2016.

\bibitem{zhu2021enriching}
Zixin Zhu, Wei Tang, Le Wang, Nanning Zheng, and Gang Hua.
\newblock Enriching local and global contexts for temporal action localization.
\newblock In {\em ICCV}, pages 13516--13525, 2021.

\end{thebibliography}
}

\end{document}